\documentclass{article}

 \usepackage[preprint]{neurips_2025}

\usepackage[utf8]{inputenc} 
\usepackage[T1]{fontenc}    
\usepackage{hyperref}       
\usepackage{url}            
\usepackage{booktabs}       
\usepackage{amsfonts}       
\usepackage{nicefrac}       
\usepackage{microtype}      
\usepackage{xcolor}         
\usepackage{graphicx}
\usepackage{amsmath}

\makeatletter
\renewcommand{\@noticestring}{%
Originally published as part of the EurIPS 2025 REO workshop [\href{https://sites.google.com/view/reoeurips?pli=1&authuser=0}{link}].%
}
\makeatother

\title{NeurEO: dissecting Earth observation embeddings with computational neuroscience}

\author{%
Thijs L van der Plas$^{1*}$ \quad Jacob JW Bakermans$^{2*}$ \quad Ioannis N Athanasiadis$^1$\\
$^1$Wageningen University \& Research, NL \quad $^2$University College London, UK \\ $^*$These authors contributed equally.\\
\texttt{\{thijs.vanderplas,ioannis.athanasiadis\}@wur.nl}, \quad 
\texttt{j.bakermans@ucl.ac.uk}
}

\begin{document}

\maketitle

\begin{abstract}
Earth embeddings are compressed representations of rich Earth observation data that enable downstream models to tackle a wide range of complex geospatial tasks. Yet, this increase in performance comes at the cost of decreased interpretability, as the semantic meaning of Earth embeddings is unknown. To address this challenge, we borrow methods from the field of computational neuroscience, which aim to interpret activity in (biological) neurons. Here, we adapt and apply these methods to assign meaning to AlphaEarth embeddings. We analyse embedding features both as stand-alone data, equivalent to spontaneous neural activity, and in relation to land cover and elevation correlates, equivalent to stimulus-response neural activity. We demonstrate correlational and multi-scale structure in the embedding features. Embedding features can display highly specific and directional tuning to land cover and elevation, such as a `North vs South woodland edge' response. When we remove the main trends of combined land cover responses (like water versus land tuning), we discover fine-grained representational details with increased effective dimensionality. Together, our analyses start to unravel what these Earth embeddings represent, which is critical for developing trustworthy downstream applications.
\end{abstract}

\section{Introduction}
Geospatial foundation models are self-supervised models that construct Earth embeddings and aim to capture rich, multi-modal Earth observation data in compressed embedding features \citep{tseng2023lightweight, brown2025alphaearth, klemmer2025satclip}. These embeddings introduce a paradigm shift for downstream geospatial tasks, presenting easily accessible, compressed representations of diverse Earth observation data on which a (simple) classifier or regression model can be trained. However, contrary to the raw Earth observation data with which they are trained, the semantic meaning of embedding representations is unknown \textit{a priori}. 
In other words, we are presented with a vast source of data, which we know should contain rich information, yet hidden to the observer. 

We argue that this scientific challenge -- interpreting what geospatial embedding features represent -- is commonly encountered in neuroscience, where recordings of neural activity are analysed to assign meaning to individual neurons or groups of neurons \citep{hubel1962receptive, o1978hippocampus, rigotti2013importance, mante2013context}. In this paper we explore this connection, by analysing Earth observation embeddings (embedding features $\times$ space) in a similar fashion to neural activity (neurons $\times$ time). Hence, where biological neural activity extends in one dimension (time), embedding features extend in two (space; latitude $\times$ longitude). In this paper we explore a variety of computational neuroscience methods originally designed to interpret neural activity, either spontaneous (\textit{i.e.}, without correlates/hypotheses) or in response to stimuli (vice versa), for both individual neurons and neuron populations, and adapt these to dissect Earth embedding features. 

\begin{figure}[t]
    \centering
    \includegraphics[width=1\linewidth]{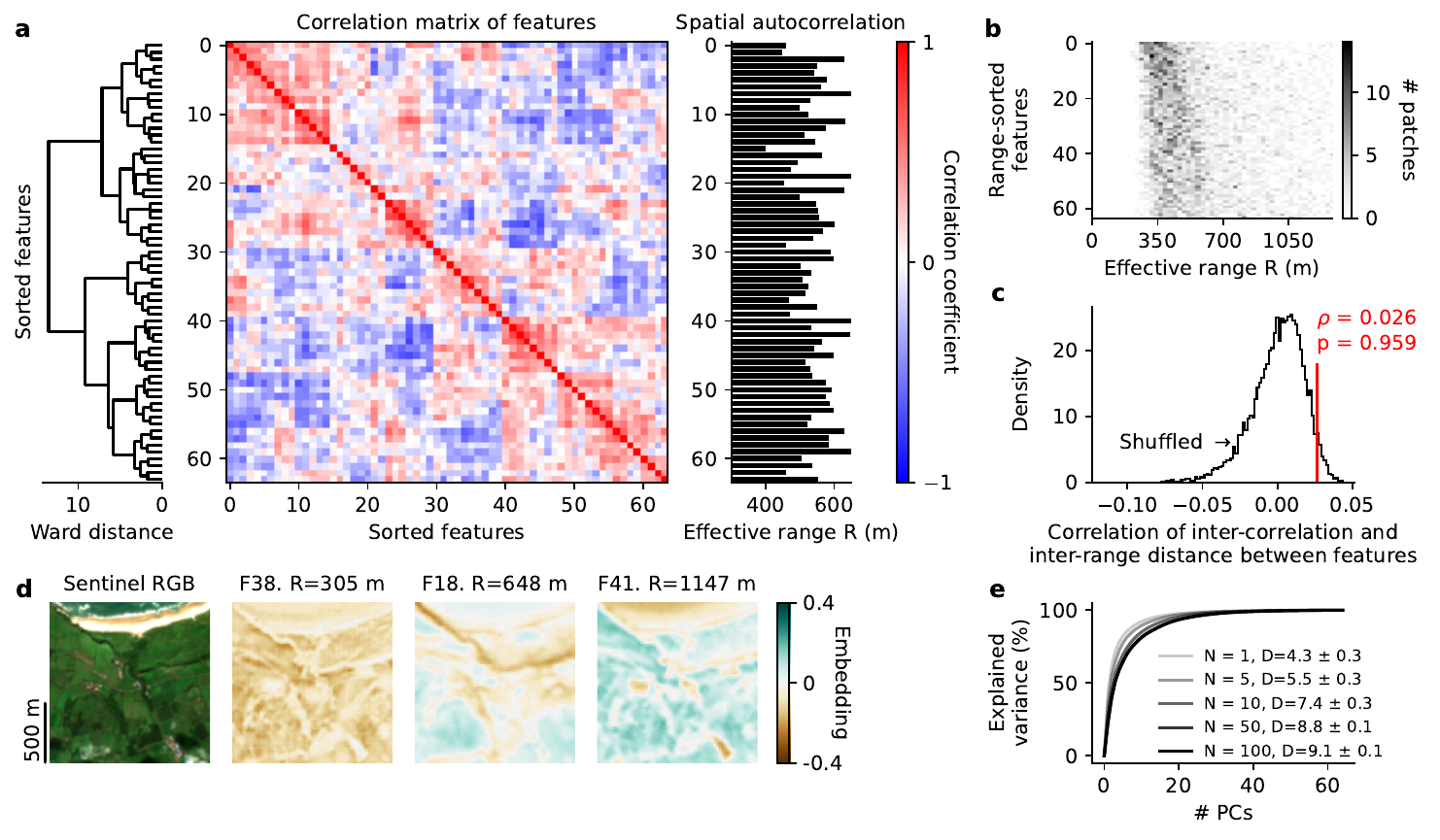}
    \caption{\textbf{Embeddings are characterised by their statistical properties.} \textbf{a)} Features exhibit a clustered (Pearson) correlation structure, sorted by hierarchical clustering (left). Median effective ranges (across patches) were calculated per feature (right). \textbf{b)} Effective range distributions vary slightly per feature (sorted in this panel only by median effective range). \textbf{c)} Variation in effective range is not correlated to variation in feature correlation structure (Pearson's $\rho=0.026$, two-sided $p = 0.082$, permutation test with $N=10,000$). \textbf{d)} Three example feature embedding maps for one example patch (F: feature, R: effective range for this patch). \textbf{e)} The effective dimensionality $D$ of features increases with increasing number of patches $N$ (mean $\pm$ standard error of the mean).}
    \label{fig:hypothesis_free}
\end{figure}

\section{Methods}
\paragraph{Dataset} A proof-of-principle dataset was constructed by randomly subsampling 100 locations from the S2BMS dataset (that contains over 1300 locations of butterfly habitats in the UK, thus representing a wide variety of UK landscapes) \citep{vanderplas2025predicting}, subsampled such that each pair of locations is at least 30 km apart. 
For each location coordinate, we downloaded 128 x 128 pixel (= 1.28 km x 1.28 km) patches of 2024 AlphaEarth embeddings \citep{brown2025alphaearth}, 2024 DynamicWorld land cover (averaged across all 2024 data points, yielding a number between 0 and 1 for each land cover class) \citep{brown2022dynamic} and Copernicus 30m DSM elevation \citep{copernicus30mDEM} data using Google Earth Engine \citep{gorelick2017google}. Additionally, Sentinel-2 RGB satellite images were downloaded for visualisation purposes only. We define the set of 9 DynamicWorld land cover classes and the DSM elevation layer as the set of 10 hypotheses to correlate embedding features with.

\paragraph{Effective range} The (variogram) effective range is defined as the inter-pixel distance where pixel values become statistically independent, and was calculated per feature per patch using a spherical fit, uniform bins and by sampling 4,000 pixels per patch with Scikit GStat \citep{malicke2022scikit}. Effective range values above the patch size 1,280 m were ignored to avoid patch edge issues, which presumably would require larger patches and sample sizes to resolve (at greater computational cost). 

\paragraph{Representational similarity analysis (RSA)} 
In neuroscience, RSA is used to test whether two separate sets of measurements (\textit{e.g.}, neural and behavioural) co-vary together \citep{kriegeskorte2008representational}.
Here, we used RSA to test whether effective range differences between features correlated significantly with pixel value correlations between features. We constructed a feature-to-feature negative distance matrix from the median effective ranges (negative distances were used such that higher values correspond to greater similarity) and a Pearson correlation matrix (of pixel values). These two matrices were correlated, and this value was compared against 10,000 permutations (where only the feature identity of the effective ranges were shuffled) to calculate a $p$ value.

\paragraph{Principal component analysis (PCA)}
In neuroscience, PCA is used to calculate the effective dimensionality of a population of neurons \citep{abbott2011interactions, farrell2022gradient}.
Here, we used PCA to calculate the effective dimensionality of the 64 embedding features for datasets of $N$ patches to quantify the redundancy among embeddings when used for a limited geographical scope (as defined by the $N$ patches). Dimensionality $D = \frac{1}{M} \sum_m^M \frac{(\sum_i \lambda_i)^2}{\sum_i \lambda_i^2}$ where $\lambda_i$ is the $i^\text{th}$ eigenvalue and $M=20$ subsamples of $N$ patches were resampled for each $N$. 
For the PCA analysis of Fig \ref{fig:hypothesis_free}e only, 165 instead of 100 S2BMS patches (at least 30 km apart) were used for subsampling. 

\begin{figure}
    \begin{minipage}{0.67\linewidth}
    \centering
    \includegraphics[width=\linewidth]{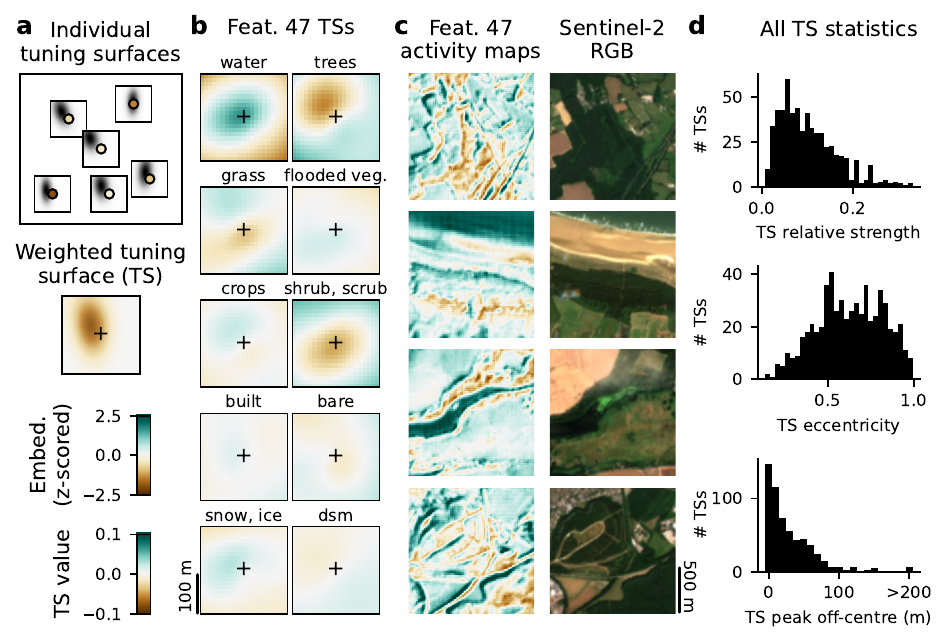}
    \end{minipage}
    \begin{minipage}{0.33\linewidth}
        \caption{\textbf{Embedding tuning surfaces reveal directional coding.} 
        \textbf{a)} Tuning surfaces (TSs) are computed by averaging hypothesis windows weighed by the focal embedding value.
        \textbf{b)} TSs of embedding feature 47, with omnidirectional coding for water and directional coding for trees.
        \textbf{c)} Embedding feature maps of feature 47 show positive coding for water bodies and positive/negative coding for Northern/Southern woodland edges.
        \textbf{d)} TS fit statistics for all feature/hypothesis combinations.
        }
        \label{fig:sta}
    \end{minipage}
\end{figure}

\paragraph{Tuning surfaces} 
In neuroscience, tuning curves/surfaces express common stimulus properties to which neurons respond, which can be calculated using the `spike-triggered-average' by temporally aligning stimulus time-series to neural `spike' times \citep{schwartz2006spike}. 
Here, we calculated a tuning surface (TS) for each feature/hypothesis combination by spatially aligning and averaging $21 \times 21$ pixel windows of hypothesis maps, each weighted by the embedding value of the central window pixel (Fig \ref{fig:sta}a). We fit resulting TSs with 2-dimensional Gaussians to obtain TS statistics across all features/hypotheses. TS relative strength was defined as the relative peak value of each hypothesis-TS per feature. TS ellipse shape was characterised by its peak-distance from (0, 0) and eccentricity $\sqrt{1 - \frac{b^2}{a^2}}$ where $a=\text{max}(\sigma_x, \sigma_y)$ and $b=\text{min}(\sigma_x, \sigma_y)$ for stds.\ $\sigma_x$ and $\sigma_y$. 


\paragraph{Canonical correlation analysis (CCA) reconstruction} 
In neuroscience, CCA is used find the linear projections of two sets of covariates that maximally correlate \citep{wang2018finding, gallego2020long}.
We used CCA to find the $N_C=10$ linear projections $\textbf{C}_F$ of the embedding feature matrix $\textbf{F}$ that maximally correlate with linear projections $\textbf{C}_H$ of the hypotheses matrix $\textbf{H}$. We then used ordinary least squares (OLS) regression to find the optimal fit of the $N_C$-dimensional $\textbf{C}_F$ components to $\textbf{F}$, which we call the reconstruction $Recon_C$ of $\textbf{F}$, and equivalently $Recon_H$ using $\textbf{H}$ to fit $\textbf{F}$ using OLS. The residuals, $Resid_C$ and $Resid_H$, are then defined by $Resid_k = \textbf{F} - Recon_k$.

\section{Results}
We use correlate-free methods to demonstrate correlational and multi-scale structure in the embeddings. We then probe the tuning of individual embedding features to land cover and elevation and find highly specific responses (\textit{e.g.}, `North vs South woodland edge'). 
We extend this tuning analysis to groups of features, which reveal fine-grained structure after the main hypothesis trends are removed.  

\paragraph{Correlate-free}
Embedding features show a strong inter-feature correlation structure, leading to clusters of similarly correlated features (Fig \ref{fig:hypothesis_free}a). 
We then used RSA to test whether clusters were predominantly caused by similarities in the spatial auto-correlation of embedding features. First, we calculated the effective range of each feature and constructed a matrix of distances between (median) effective ranges of all feature pairs. We then tested whether this matrix was significantly correlated to the inter-feature correlation structure, but found that is was very weakly correlated ($\rho=0.026$, two-sided $p=0.082$, Fig \ref{fig:hypothesis_free}b-d). Thus, similar features may have similar scales, but there is also considerable variation of scale within clusters. Further, the PCA dimensionality increased with the number of patches (Fig \ref{fig:hypothesis_free}e), demonstrating that diversifying the dataset can decorrelate the features.

\paragraph{Individual tuning}
We computed tuning surfaces for all combinations of embedding features and hypotheses to determine if any features directionally encode any hypotheses (Fig \ref{fig:sta}a). Indeed, we found a number of such examples, for example feature 47, which responds to water bodies and, specifically, distinguishes between Northern and Southern woodland edges (Fig \ref{fig:sta}b, c). The selectivity of embedding features to specific hypotheses varied, as defined by their relative tuning strength, and furthermore we observed that off-centre peaks can be used to identify directional coding (Fig \ref{fig:sta}b, d).

\begin{figure}
    \begin{minipage}{0.67\linewidth}
    \centering
    \includegraphics[width=\linewidth]{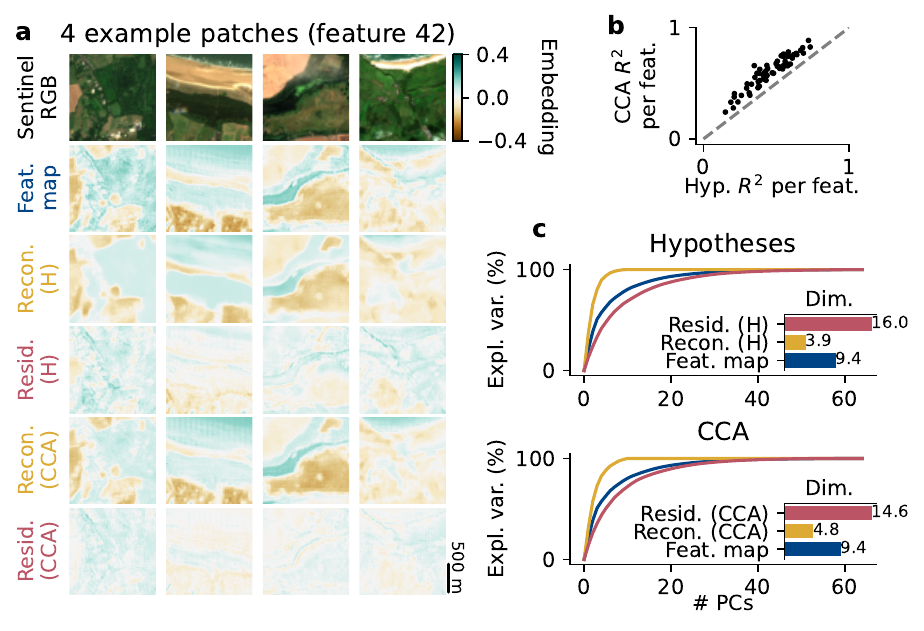}
    \end{minipage}
    \begin{minipage}{0.33\linewidth}
        \caption{\textbf{Embeddings can be linearly split into low-dimensional reconstructions and high-dimensional residuals.} 
        \textbf{a)} RGB, embedding feature maps, reconstructions and residuals for hypotheses (H) and CCA for an example embedding feature with $R^2_H=0.72$ and $R^2_{CCA}=0.88$.
        \textbf{b)} CCA reconstructions explain more embedding variance ($R^2$) than hypotheses reconstructions.
        \textbf{c)} PCA dimensionality (dim.) is greatest for residuals, followed by embeddings and reconstructions. 
        }
        \label{fig:cca}
    \end{minipage}
\end{figure}

\paragraph{Population coding}
Finally, we asked how much variance of the embedding features is explained by the land cover and elevation hypotheses. Using OLS regression, we found that the 10 hypotheses explain on average $R^2 =0.46$ of embedding variance. Next, because combinations of embedding features might represent combinations of hypotheses, we used canonical correlation analysis (CCA) to find the embedding reprojections that maximally correlate to the hypotheses, which explained $R^2=0.60$ of variance on average (Fig \ref{fig:cca}a,b). Both residuals (\textit{i.e.}, the embeddings after subtracting the regression fits) reveal new details, as the correlations with the land cover and elevation hypotheses have effectively been removed (Fig \ref{fig:cca}a). Accordingly, the effective dimensionality of residuals is greater than the original embedding dimensionality (Fig \ref{fig:cca}c).

\section{Discussion}
\paragraph{Limitations} 
We aimed to demonstrate methods new to Earth observation data analysis that can be helpful for interpreting Earth embeddings, using a small, proof-of-concept dataset of AlphaEarth embeddings.
A larger and more diverse dataset is needed to reach  more definitive conclusions about AlphaEarth embedding features, and other Earth embeddings should be considered to achieve more generalised findings. It would be particularly interesting to compare Earth embeddings trained with different model architectures and loss functions, such as Presto \citep{tseng2023lightweight}, SatCLIP \citep{klemmer2025satclip} and spherical harmonics \citep{russwurm2023geographic}. 
\paragraph{Conclusions}  
Earth embeddings are highly compressed, yet for local applications (that do not encompass the global variety of land surfaces), these embeddings might still suffer from redundancies. We show that for a diverse dataset of locations in the United Kingdom, the effective dimensionality converges to $< 10$, suggesting that a more efficient embedding representation may be derived for \textit{any} localised application. 
We further show that regressing out the projections maximally correlated to land cover and elevation increases the dimensionality of the residual embedding features, revealing new spatial variability that used to be obscured by large variance modes. Together, these findings suggest that Earth embeddings can be fine-tuned for local applications by reprojecting, selecting and assigning semantic meaning to embedding features, originally designed to describe the entire globe. 

\begin{ack}
TLvdP and INA acknowledge funding from the Dutch Research Council (NWO) Large-Scale Research Infrastructures (LSRI) programme for the LTER-LIFE (\url{http://www.lter-life.nl}) infrastructure (grant 184.036.014). JJWB acknowledges funding from the Royal Society through a Royal Society Newton International Fellowship (NIF\textbackslash R1\textbackslash 252778). The authors thank Marc Ru{\ss}wurm for constructive feedback and suggestions.
\end{ack}

\bibliographystyle{plainnat}
\bibliography{neureo_bib}

\end{document}